\documentclass[journal,12pt,onecolumn,draftclsnofoot,]{IEEEtran}
\usepackage{times,amsmath,epsfig}
\usepackage{relsize}
\usepackage{amsmath}
\usepackage{fixltx2e}
\usepackage{graphicx}
\usepackage{lineno,hyperref}
\usepackage{tabularx}
\usepackage{epstopdf}
\usepackage{pgfplots}
\usepackage{caption}
\usepackage{rotating}
\usepackage{csquotes}
\usepackage{chngpage}
\renewcommand{\figurename}{Fig.}
\modulolinenumbers[5]
\title{Content-Based Image Retrieval Based on Late Fusion of Binary and Local Descriptors}
\author{
    \IEEEauthorblockN{Nouman Ali\IEEEauthorrefmark{1,2}, Danish Ali Mazhar\IEEEauthorrefmark{2}, Zeshan Iqbal\IEEEauthorrefmark{2}, Rehan Ashraf\IEEEauthorrefmark{1},Jawad Ahmed\IEEEauthorrefmark{3}, Farrukh Zeeshan Khan\IEEEauthorrefmark{2}} \\
    \IEEEauthorblockA{\IEEEauthorrefmark{1}Department of Computer Science, National Textile University, Faisalabad, Pakistan
  }
   \\ \IEEEauthorblockA{\IEEEauthorrefmark{2}Faculty of Telecommunication \& Information Engineering, University of Engineering \& Technology, Taxila, Pakistan
   }
   \\ \IEEEauthorblockA{\IEEEauthorrefmark{3}Department of Basic Sciences, University of Engineering \& Technology, Taxila, Pakistan
   }
  \\ \{noumanali\}@ntu.edu.pk
}

\begin{document}
\maketitle
\thispagestyle{empty}
\pagestyle{empty}

\begin{abstract}
One of the challenges in Content-Based Image Retrieval (CBIR) is to reduce the semantic gaps between low-level features and high-level semantic concepts. In CBIR, the images are represented in the feature space and the performance of CBIR depends on the type of selected feature representation. Late fusion also known as visual words integration is applied to enhance the performance of image retrieval.  The recent advances in image retrieval diverted the focus of research towards the use of binary descriptors as they are reported computationally efficient. In this paper, we aim to investigate the late fusion of Fast Retina Keypoint (FREAK) and Scale Invariant Feature Transform (SIFT). The late fusion of binary and local descriptor is selected because among binary descriptors, FREAK has shown good results in classification-based problems while SIFT is robust to translation, scaling, rotation and small distortions. The late fusion of FREAK and SIFT integrates the performance of both feature descriptors for an effective image retrieval. Experimental results and comparisons show that the proposed late fusion enhances the performances of image retrieval.
\end{abstract}

%
\section{Introduction}
In image classification and retrieval-based problems, the extraction of a meaningful image descriptor is an open research problem \cite{dharani2013survey}. Due to an increase in the number of image archives, it is necessary to design an effective system for image search \cite{dharani2013survey,alzu2015semantic}. The conventional annotations-based approach for image retrieval relies on text and keywords-based image search \cite{dharani2013survey,kumar2013content,rashedi2013simultaneous}. Human manual efforts and difference in the visual perception make the conventional approach less effective \cite{dharani2013survey,alzu2015semantic}. The modern image search systems retrieve the images on the basis of image visual contents and this is referred as CBIR \cite{dharani2013survey}. Image retrieval has vast applications in many domains like image analysis, search of image over internet, medical image retrieval, remote sensing and video surveillance \cite{dharani2013survey}. The recent research is carried out to enhance the performance of image retrieval by retrieving the images that are similar to the query image \cite{dharani2013survey,zhang2012review,ali2016image}. In recent few years, different types of image representations are proposed that are associated with different application domains \cite{dharani2013survey,alzu2015semantic,mukherjee2015comparative,ashraf2014novel,ashraf2015content,ashraf2016content}. In this paper, we are interested to explore a novel image representation based on the late fusion of binary and local features.  The proposed framework can describe the visual contents in a meaningful way and the main focus of this research is image retrieval.

In CBIR, the commonly used image representations are mainly based on global and local features \cite{dharani2013survey,alzu2015semantic,mehmood2016novel}. The global feature representations are based on color, texture and shape \cite{dharani2013survey}. Color is one of the fundamental image feature and it is not dependent on size, direction and angle \cite{alzu2015semantic}. Color features lack spatial distribution and perceptual meaning \cite{dharani2013survey,alzu2015semantic}. Texture features are divided into two main classes and texture is extracted to capture the spatial attributes in the group of pixels \cite{alzu2015semantic,zhang2012review}. The main limitation of spatial texture techniques is their sensitivity to noise and distortion \cite{alzu2015semantic,zhang2012review}. The spectral texture techniques are dependant on shape and have shown good performance in-case of square regions \cite{zhang2012review}. The sensitivity to noise and computational complexity limits the use of texture features in many application domains \cite{alzu2015semantic,zhang2012review}. According to Zhang et al. \cite{zhang2004review}, region-based and contour-based are the two main approaches that are commonly applied to extract shape features. The sensitivity to scaling, translation and rotation are the main limitations of shape-based image representations \cite{alzu2015semantic,zhang2012review}.

The local feature descriptors like Scale Invariant Feature Transform (SIFT) \cite{lowe2004distinctive}, Speeded-Up Robust Features (SURF) \cite{bay2006surf} and Histogram of Oriented Gradients (HOG) \cite{dalal2005histograms} have shown good performance in various applications of image retrieval \cite{mukherjee2015comparative,krajnik2015image}. Local feature descriptors are robust to translations, scaling, rotation, and small distortions and are applied in various image matching applications \cite{mukherjee2015comparative}. According to the recent literature \cite{mukherjee2015comparative}, the classification and retrieval performance of local features is better than global features.

The recent advances in computer vision turned the attention of research to the use of binary descriptors \cite{mukherjee2015comparative,figat2014performance}. Binary Robust Invariant Scalable Keypoints (BRISK) \cite{leutenegger2011brisk}, Fast Retina Keypoint (FREAK) \cite{alahi2012freak}, Binary Robust Independent Elementary Features (BRIEF)\cite {calonder2010brief} and Features from Accelerated Segment Test (FAST) \cite{viswanathan2009features} are the examples of binary descriptors. Local and binary descriptors are used to describe the image information in a high-dimensional feature space by using keypoints \cite{mukherjee2015comparative}. The late fusion (also known as visual word integration) based on proper features selection improves the effectiveness of image retrieval. \cite{zhang2012review,yu2013feature,yuan2011sift}. The performance of FREAK is reported efficient for classification-based problems \cite{mukherjee2015comparative,figat2014performance} while SIFT is robust to translation, scaling, rotation, and small distortions \cite{valgren2010sift}.

Due to these facts, in this paper, we investigated the late fusion/visual word integration of FREAK and SIFT. Binary and local features are extracted from training images. A quanization algorithm such as \textit{k}-means is applied to cluster two codebooks (visual vocabularies), that are based on descriptors of FREAK and SIFT, respectively. The codebooks of binary and local features are combined to represent an image in a dimension that is twice the size of constructed vocabulary. The main contributions of the proposed research is the CBIR based on the late fusion of FREAK (binary descriptor) and SIFT (local descriptor).

The remaining paper is organized as follow: Section \ref{sec:2} describes the related work. Section \ref{sec:3} provides an overview about the proposed research methodology. Section \ref{sec:4} presents evaluation measures and results obtained on three image benchmarks while Section \ref{sec:5} concludes our work and points towards the future directions in the research.

\section{Related Work}
\label{sec:2}

CBIR is used to search the images that are  visually similar to the query image \cite{dharani2013survey,ashraf2016content,localglobal}. Wang et al. \cite{wang2013spatial} proposed Spatial Weighting BOF (SWBOF) by extracting the spatial information from different sub-blocks of the images. Adjacent block distance, local entropy and variance are used for the extraction of spatial information. The spatial weighting is achieved by weighting the correspondence visual words with the help of texture information. Yuan et al. \cite{yuan2011sift} proposed image retrieval by using integration SIFT and Local Binary Pattern (LBP). The combination of SIFT-LBP is selected to enhance the performance of image retrieval in-case of noisy background and ambiguous objects.  Yu et al. \cite{yu2013feature} proposed two new image representations based on early features fusion for an effective image retrieval. The clusters are constructed by applying a weighted scheme to maintain a balance between two features. The two new feature integrations are image-based SIFT-LBP and patch-based HOG-LBP. The image-based integration of SIFT-LBP outperforms the state-of-the-art approaches \cite{yu2013feature}.  Zhang et al. \cite{zheng2010image} proposed a rotation invariant image matching system by using combination of SIFT and LBP. The regions of LBP descriptors are calculated by using SIFT detector. Kabbai et al. \cite{kabbai2015image} proposed a new approach to extract invariant features from the regions of interest. The uniform pattern is applied to the LBP and Center Symmetric Local Binary Pattern (CSLBP) for a robust image matching application.

 According to \cite{ali2016image}, the spatial information can be extracted by dividing an image into Level 1 and Level 2 triangles. Dense SIFT is used for feature extraction and three different classifiers are applied to determine the best retrieval performance. Zeng et al. \cite{zeng2016image} proposed an image representation that is based on generalized histogram of quantized colors. Gaussian Mixture Models (GMMs) is applied for quantization and Expectation-Maximization (EM) algorithm is used for training. Bayesian Information Criterion (BIC) is applied for determination of quantized color bins. Images are retrieved on the basis of similarity between the respective spatiograms. Walia et al. \cite{walia2014fusion} proposed a fusion framework for color-based image retrieval. The color and texture are extracted by applying Color Difference Histogram (CDH) and Angular Radial Transform (ART) and modification in CDH algorithm is proposed to make it more effective.  Dubey et al. \cite{dubey2015rotation} proposed a rotation and scale invariant hybrid image descriptor for an efficient image retrieval. The color features are extracted by quantizing RGB color space while texture is extracted by structuring the patterns that are generated from locally structured elements. Color and textural are integrated to construct the inherently Rotation and Scale-invariant Hybrid image Descriptor (RSHD). Montazer et al. \cite{montazer2015improved} proposed a new learning method for RBFNNs (Radial Basis Function Neural Networks). Particle Swarm Optimization (PSO) is applied to initialize the radial basis function units in more accurate way. The spatial information of the data and non-linearity of the function are approximated to determine the widths of RBFNNs. According to Wan et al. \cite{wan2014deep}, the modern machine learning techniques based on Convolutional Neural Networks (CNN) can reduce the semantic gaps. CNN model pre-trained on large dataset can be used for feature extraction and the optimized feature extraction techniques based on deep learning outperforms conventional features. In our previous work \cite{ali2016novel}, we proposed the visual words integration (late fusion) of two local features. Different Weighed Averages (WA) of local features are also calculated to sort out the second best performance for image retrieval. The research presented in this paper is different from our previous work \cite{ali2016novel} as in this paper, we replaced SURF with FREAK to evaluate the late fusion (visual words integration) of SIFT (local descriptor) with FREAK (binary descriptor). The experimental results demonstrates the dominant effect of FREAK when used with the late fusion SIFT descriptor.
 \section{Bag of Features Representation }
\label{sec:3}
\begin{figure}[!h]
\centering
    \includegraphics[width=0.99\textwidth, height=10cm]{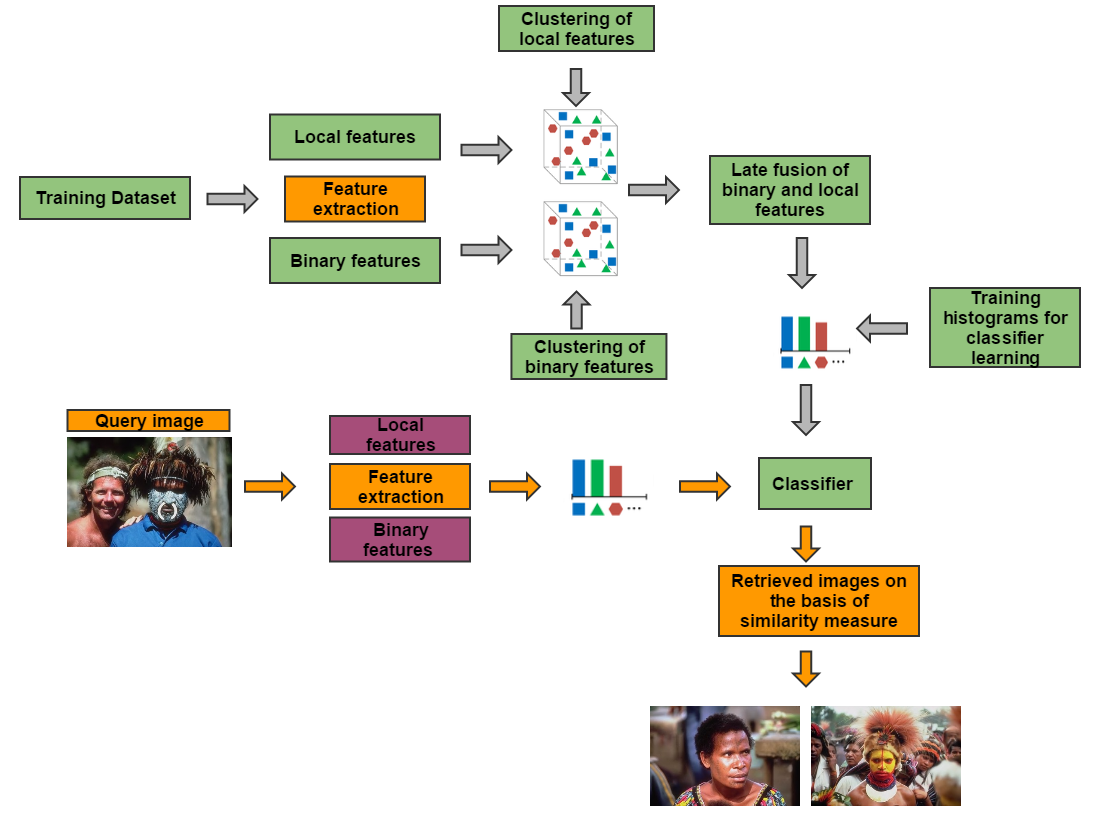}
      \caption{Proposed framework based on late fusion of FREAK and SIFT.}\label{fig2}
\end{figure}

Fig. \ref{fig2} represents the block diagram of the proposed image representation that is based on late fusion of binary and local features. The proposed research is based on BoVW image representation model \cite{sivic2003video}. The detail about features extraction using FREAK, SIFT and late fusion of FREAK and SIFT is mentioned in the following sub-sections.
\subsection{Fast Retina Keypoint (FREAK)}\label{3.1}

FREAK is a binary descriptor that is computed on the basis of brightness comparison tests around the keypoints and is inspired from the human visual system \cite{alahi2012freak}. In the first step, a circular sampling grid is applied to generate retinal sampling pattern with higher density of points near the center . One-bit Difference of Gaussian (DoG) is applied to compute a binary descriptor. The feature are calculated by applying a saccadic search. In the last step, the sum of local gradients over selected pairs are used to compute the rotation of keypoints \cite{alahi2012freak}.

\subsection{Scale Invariant Feature Transform (SIFT)}\label{3.1}
There are four main steps that are involved to compute SIFT descriptors \cite{lowe2004distinctive}. The first step involves the computation of interest points, for this purpose the Difference of Gaussian (DoG) is used to produce several Gaussian blurred images. The neighborhood image are compared for the calculation of DoG. The second step involves the calculation of extrema to find out the stable keypoints and low contrast and pints along the edges are removed by apply Taylor series.  Determinant and trace of Hessian metric are used to remove the outliers. In the third step, to achieve rotation invariance, the principal orientation is calculated for the keypoints. The last step involves the computation of SIFT descriptor. For each keypoint, a set of orientations histograms are created on 4x4 pixel neighborhoods, with 8 orientation bins \cite{lowe2004distinctive}.

\subsection{Proposed Late Fusion Based on Binary and Local Descriptors}\label{3.3}
\begin{enumerate}
  \item  In BoVW model, a raw image I is represented as:
  \begin{equation}
I=(a_{ r,s})
\end{equation}
\begin{center}
where {\textit{r,s}} are the pixels of image.
\end{center}
\item Binary and local features (FREAK and SIFT) are extracted from a set of training images and an image {\textit{I}} is represented as:
\begin{equation}
I= \lbrace d_{1},d_{2},....,d_{T} \rbrace
\end{equation}
\begin{center}
Where {\textit{d\textsubscript{1}}} to {\textit{d\textsubscript{T}}} are the image descriptors. \newline
\end{center}
\item A quantization algorithm such as {\textit{k}}-means is applied to construct the codebook (visual vocabulary) consisting of Z words, represented as {\textit{CB:}}
\begin{equation}
CB=\lbrace w_{1},w_{2},....,w_{Z} \rbrace
\end{equation}
where {\textit{CB}} is the codebook consisting of  {\textit{w\textsubscript{Z}}} visual words, separate codebooks are constructed for FREAK and SIFT by extracting the respective features.
\hfill \break
\item The features are extracted from the image and quantized in the feature space. Mapping of each visual word is done over the image areas and the nearest words are assigned to the quantized descriptors according to the following equation:
\newline
\begin{equation}
w(d_{m})=\underset{w\;\varepsilon\;CB}{\text{argmin}}\;Dist(w,d_{m})
\end{equation}
where {\textit{w(d\textsubscript{m})}} is representing the visual word assigned to the{\textit{ m\textsuperscript{th}}} descriptor {\textit{d\textsubscript{m}}} while {\textit{Dist(w,d\textsubscript{m})}} is the distance between the descriptor {\textit{d\textsubscript{m}}} and visual word {\textit{w}}. Each image is represented as a collection of patches and each patch is represented by a visual word (visual words of FREAK are mapped on the images by using the codebook of FREAK while visual words of SIFT are mapped on the images by the using the codebook of SIFT).\newline

\item The histograms consisting of the visual words of FREAK and SIFT are combined and an image is represented in a dimension that is twice the size of constructed codebook. Consider Z as the number of visual words of the codebook. Let {\textit{F\textsubscript{i}}} be the set of the descriptors that are mapped to the visual word {\textit{w\textsubscript{i}}} then the {\textit{i\textsuperscript{th}}} bin of the histogram of visual words {\textit{b\textsubscript{i}}}, is the cardinality of the set {\textit{D\textsubscript{i}}}.
\begin{equation}
b_{i}=Card(F_{i})\quad and \quad F_{i}=\lbrace d_{m},m \in(1,....,Z)\mid w(d_{m})=w_{i} \rbrace
\end{equation}

\end{enumerate}

\section{Experimental Parameters and Results}
\label{sec:4}

The proposed late fusion is evaluated by using three image dataset \cite{wang2001simplicity,li2008real,oliva2001modeling}. We used the image classification parameters as mentioned in \cite{ali2016novel}. 70\% of the images from each class are selected for training and 30\% for testing. Keeping in view, the unsupervised nature of clustering using \textit{k}-means, each experiment is repeated 10 times. During every run, images are randomly selected for training and testing. The set of selected images for training are used for the construction of codebook and image retrieval performance is reported by using the images from test dataset. The images are retrieved by calculating the closeness among the classifier score values within the same class.

The size of the codebook affects image retrieval performance \cite{nowak2006sampling,csurka2004visual}. An increase in the the size of the codebook increases the retrieval precision. The larger size codebook decreases the retrieval precision due to over-fitting \cite{nowak2006sampling,csurka2004visual}. We constructed different sizes of codebook from randomly set of selected training images to sort out the best retrieval performance. The codebook is constructed by a random selection of 25 \%, 50 \% and 75 \% of features (both for FREAK and SIFT) per image (from the training dataset). Additional experiments are performed to evaluate the performance of FREAK and SIFT separately with the same experimental parameters to  represent the dominant affect of late fusion on image retrieval performance.

\subsection{Evaluation Measures}

We selected precision and recall to determine the performance of our proposed late fusion \cite{alzu2015semantic}. Precision determines the number of correctly retrieved images. \hfill \break
\begin{equation}
Precision=\frac{K_{r}}{X{r}}
\end{equation}

where K\textsubscript{r} represents the number of relevant images similar to the query and X\textsubscript{r} indicates the number of images retrieved by the system in response to the query.
 \hfill \break
\begin{equation}
Recall=\frac{K_{r}}{X{c}}
\end{equation}
where X\textsubscript{c} is total number of images of that class in the database.

\begin{figure}[!h]
\centering
    \includegraphics[width=0.90\textwidth, height=3.5cm]{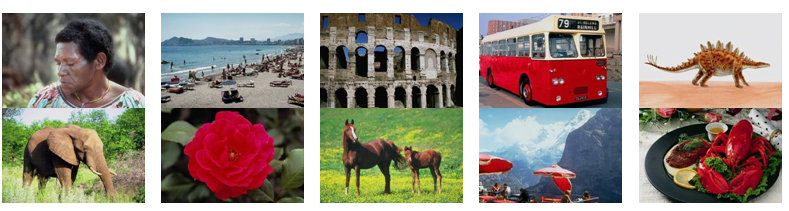}
      \caption{Randomly selected images from each class of Corel-1000 image dataset \cite{wang2001simplicity}. }\label{figcorel}
\end{figure}
\subsection{Performance Evaluation using Corel-1000}

There are 10 classes in Corel-1000 image dataset and each class contains 100 images. Fig. \ref{figcorel} represents a sample of randomly selected images from each class of the Corel-1000. We selected Corel-1000 for the evaluation of proposed framework as it has been recently used to evaluate the performance CBIR research \cite{yu2013feature,wang2013spatial,ali2016novel,montazer2015improved,tian2014feature}. We varied the codebook size to sort out the best retrieval performance of proposed work. Table \ref{tab:Table1} presents the numerical values of Mean Average Precision (MAP) obtained from the proposed framework for top 20 retrievals. Fig. \ref{fig:3} presents a comparison of MAP as a function of codebook size using different evaluation parameters.

\begin{table}[!h]
\centering
\caption{ MAP as a function of codebook size (late fusion of FREAK and SIFT)}.
\begin{tabular}
{|p{2.2cm}|>{\centering}p{0.65cm}|>{\centering}p{0.65cm}|>{\centering}p{0.65cm}|>{\centering}p{0.65cm}|>{\centering}p{0.75cm}|>{\centering}p{0.75cm}|>{\centering}p{0.65cm}|}
\hline
Codebook size and features \% used & 50 & 100 & 200 & 300 &
400 &600 &800
 \tabularnewline \hline
\centering 25 \% &    68.85 & 70.45 & 74.36 & 74.52 & 74.68 & 74.58 & 73.99
 \tabularnewline \hline
 \centering 50 \% & 69.31 & 70.56 & 73.86 & 74.91 & 74.16 & 74.96 & 74.01
 \tabularnewline \hline
 \centering 75 \% & 70.45 & 71.13 & 73.99 & 73.69 & 74.55 & 74.87 & 74.25
 \tabularnewline \hline
\centering \textbf{Mean}  & 69.54 & 70.71 & 74.07 & 74.37 & \textbf{74.46} & \textbf{74.80} & 74.08 \tabularnewline \hline

\end{tabular}\label{tab:Table1}
\end{table}

\begin{center}
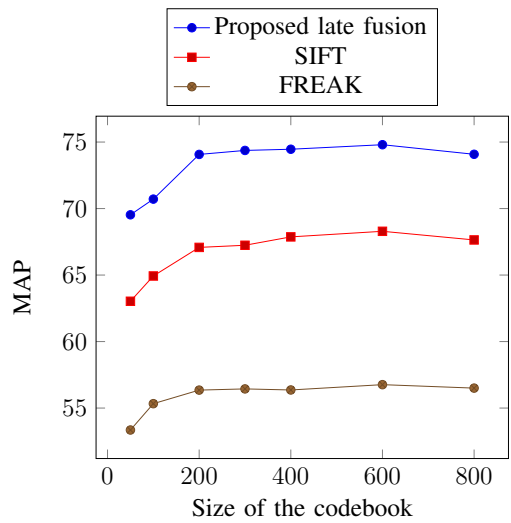

\begin{tikzpicture}[scale=0.80]

\pgfplotsset{every axis legend/.append style={
		at={(0.5,1.03)},
		anchor=south}}
\begin{axis}[legend columns=1,ylabel=MAP ,xlabel=Size of the codebook]
\addplot coordinates { (50,69.53) (100,70.71)  (200,74.07)(300,74.37) (400,74.46)(600,74.80)(800,74.08)};

\addplot coordinates {(50,63.03) (100,64.93)  (200,67.08)(300,67.24) (400,67.87)(600,68.29)(800,67.64) };

\addplot coordinates {(50,53.35) (100,55.33)  (200,56.35)(300,56.44) (400,56.36)(600,56.76)(800,56.50)};

\addlegendentry{Proposed late fusion}

\addlegendentry{SIFT}

\addlegendentry{FREAK}

 \end{axis}
   \end{tikzpicture}
   \captionof{figure}{MAP as a function of codebook size (Corel-100).} \label{fig:3}
\end{center}

The results and comparisons presented in Table \ref{tab:Table1} and Fig. \ref{fig:3} indicate that the best MAP is obtained from the proposed last fusion on a codebook of size 600 with a value of 74.80 \%. The best MAP obtained by using SIFT and FREAK on a codebook with a size of 600 is 68.29\% and 56.76\%, respectively. The MAP on codebook of each size by using the proposed late fusion is higher than that of SIFT and FREAK.
To present a sustainable performance, the best MAP obtained from the proposed late fusion is compared with existing research \cite{yu2013feature,wang2013spatial,ali2016novel,montazer2015improved,tian2014feature}. Table \ref{tab:Table2} and Table \ref{tab:Table3} presents the comparison of precision and recall for top 20 retrievals ( as the research selected for comparison is also reported for top 20 retrievals).

\begin{table}[!h]
\small
\centering
\caption{Comparison of retrieval precision.}
\begin{tabular}
{|p{1.6cm}|>{\centering}p{1.3cm}|>{\centering}p{1.42cm}|>{\centering}p{1.1cm}|>{\centering}p{1.0cm}|>{\centering}p{1.0cm}|>{\centering}p{1.0cm}|}
\hline
Name of class and method & Proposed framework &WA SIFT - SURF \cite{ali2016novel} & RBF-NN \cite{montazer2015improved}  & Color SIFT \cite{tian2014feature} &
Spatial SIFT \cite{wang2013spatial} & SIFT-LBP  \cite{yu2013feature}
 \tabularnewline  \hline
 Africa & 63.64 & \textbf{69.75} & 58.73 & \textbf{74.60} & 64    & 57
   \tabularnewline  \hline
 Beach & \textbf{60.99} & 54.25 & 48.94 & 37.80 & 54    & \textbf{58}
  \tabularnewline \hline
 Buildings & \textbf{68.21} & \textbf{63.95} & 53.74 & 53.90 & 53    & 43
 \tabularnewline \hline
 Buses & 92.75 & 89.65 & \textbf{95.81} & \textbf{96.70} & 94    & 93
   \tabularnewline \hline
   Dinosaurs & \textbf{100.00} & 98.7  & 98.36 & \textbf{99} & 98    & 98
    \tabularnewline \hline
Elephants & \textbf{72.64} & 48.8  & 64.14 & 65.90 & \textbf{78} & 58
   \tabularnewline \hline
Flowers & \textbf{91.54} & \textbf{92.3} & 85.64 & 91.20 & 71    & 83
    \tabularnewline \hline
Horses & 80.06 & \textbf{89.45} & 80.31 & \textbf{86.90} & 93    & 68
   \tabularnewline \hline
   Mountains &  \textbf{59.67} & 47.3  & 54.27 & \textbf{58.50} & 42    & 46
    \tabularnewline \hline
 Food & 58.56 & \textbf{70.9} & \textbf{63.14} & 62.20 & 50    & 53
   \tabularnewline \hline
  \textbf{Mean} & \textbf{74.80} & \textbf{70.58} & 70.31 & 72.67 & 69.70 & 65.70
   \tabularnewline \hline
\end{tabular}
\label{tab:Table2}
\end{table}
\begin{table}[!h]
\small
\centering
\caption{Comparison of recall.}
\begin{tabular}
{|p{1.6cm}|>{\centering}p{1.3cm}|>{\centering}p{1.42cm}|>{\centering}p{1.1cm}|>{\centering}p{1.0cm}|>{\centering}p{1.0cm}|>{\centering}p{1.0cm}|}
\hline
Name of class and method & Proposed framework &WA SIFT - SURF \cite{ali2016novel} & RBF-NN \cite{montazer2015improved}  & Color SIFT \cite{tian2014feature} &
Spatial SIFT \cite{wang2013spatial} & SIFT-LBP  \cite{yu2013feature}
 \tabularnewline  \hline
 Africa & 12.73 & \textbf{13.95} & 11.75 & \textbf{14.92} & 12.80 & 11.40
   \tabularnewline  \hline
 Beach & \textbf{12.20} & 10.85 & 9.79  & 7.56  & 10.80 & \textbf{11.60}
  \tabularnewline \hline
 Buildings & \textbf{13.64} & \textbf{12.79} & 10.75 & 10.78 & 10.60 & 8.60
 \tabularnewline \hline
 Buses &  18.55 & 17.93 & \textbf{19.16} & \textbf{19.34} & 18.80 & 18.60
   \tabularnewline \hline
   Dinosaurs & \textbf{20.00} & 19.74 & 19.67 & \textbf{19.80} & 19.60 & 19.60
    \tabularnewline \hline
Elephants & \textbf{14.53} & 9.76  & 12.83 & 13.18 & \textbf{15.60} & 11.60
   \tabularnewline \hline
Flowers & \textbf{18.31} & \textbf{18.46} & 17.13 & 18.24 & 14.20 & 16.60
    \tabularnewline \hline
Horses & 16.01 & \textbf{17.89} & 16.06 & \textbf{17.38} & 18.60 & 13.60
   \tabularnewline \hline
   Mountains & \textbf{11.93} & 9.46  & 10.85 & \textbf{11.70} & 8.40  & 9.20
    \tabularnewline \hline
 Food & 11.71 & \textbf{14.18} & \textbf{12.63} & 12.44 & 10.00 & 10.60
   \tabularnewline \hline
 \textbf{Mean} &\textbf{14.96} & \textbf{14.12} & 14.06 & 14.53 & 13.94 & 13.14
   \tabularnewline \hline
\end{tabular}
\label{tab:Table3}
\end{table}

\begin{center}
\begin{figure}[!h]
\centering
    \includegraphics[width=0.90\textwidth,height=8cm]{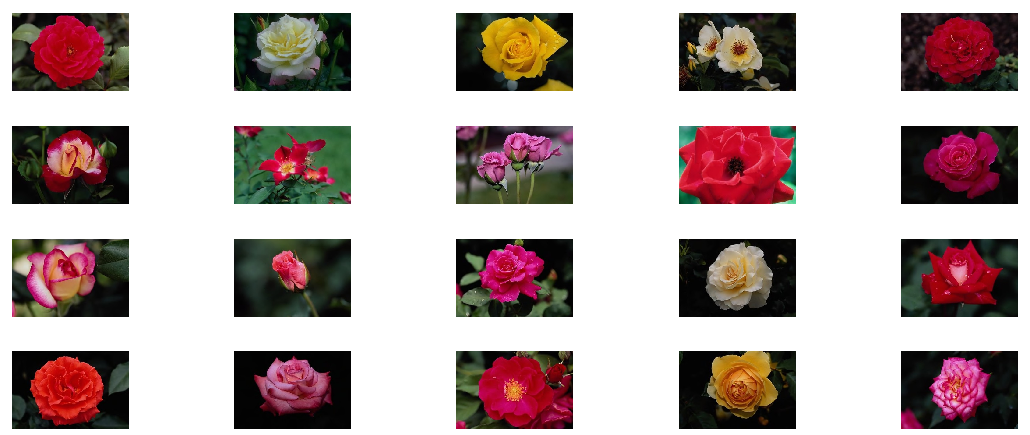}
        \caption{Retrieval result on the basis of similarity measures for the semantic class "Flowers"}\label{fig:4}
\end{figure}
\end{center}

\begin{center}
\begin{figure}[!h]
\centering
    \includegraphics[width=0.90\textwidth,height=6cm]{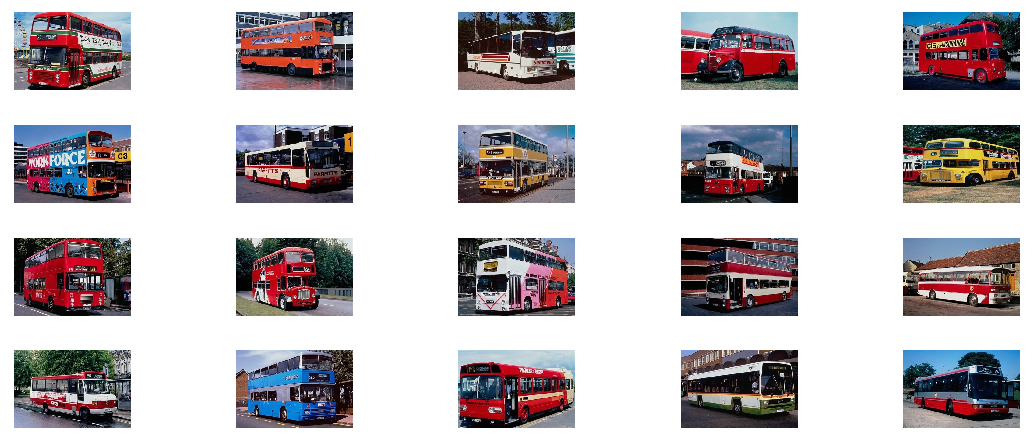}
        \caption{Retrieval result on the basis of similarity measures for the semantic class "Buses"}\label{fig:5}
\end{figure}
\end{center}

The best values of precision and recall are mentioned as bold in Table \ref{tab:Table2} and Table \ref{tab:Table3}. The MAP of proposed late fusion outperforms state-of-the-art research \cite{yu2013feature,wang2013spatial,ali2016novel,montazer2015improved,tian2014feature}.
Fig. \ref{fig:4} and Fig. \ref{fig:5} represents the image retrieval results obtained on the basis of similarity measures for the semantic class "Flowers" and "Buses", respectively. The classifier output label determines the class of the image while the images are retrieved on the basis of closeness among classifier score values.

\begin{figure}[!h]
\centering
    \includegraphics[width=0.90\textwidth, height=4cm]{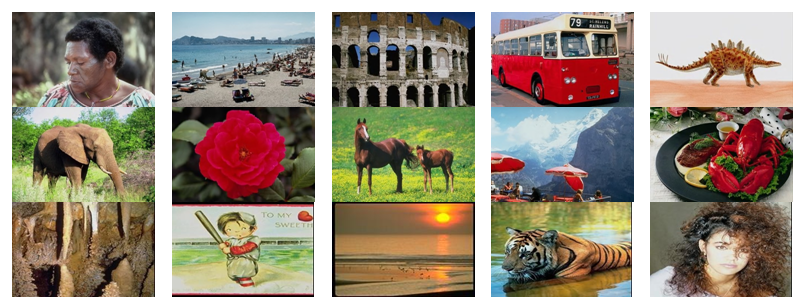}
      \caption{Randomly selected images from each class of Corel-1500 image dataset \cite{li2008real}.}\label{figcorel1500}
\end{figure}
\subsection{Performance Evaluation Using Corel-1500 }

There are 15 classes in the Corel-1500 \cite{li2008real} image benchmark and each class contains 100 images \cite{li2008real}. Fig. \ref{figcorel1500} represents a sample of randomly selected images from each class of the Corel-1500. We evaluated the proposed framework on Corel-1500 and compared the results with state-of-the-art CBIR methods \cite{zeng2016image,ali2016novel}. Fig. \ref{fig:3} presents a comparison of MAP as a function of codebook size using different evaluation parameters.

\begin{center}
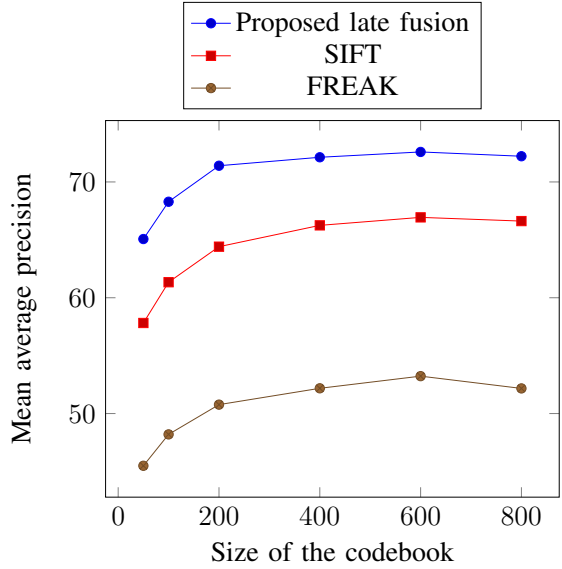

\begin{tikzpicture}[scale=0.88]

\pgfplotsset{every axis legend/.append style={
		at={(0.5,1.03)},
		anchor=south}}
\begin{axis}[legend columns=1,ylabel=Mean average precision,xlabel=Size of the codebook]
\addplot coordinates { (50,65.07) (100,68.29)  (200,71.40)(400,72.13) (600,72.59)(800,72.22)};

\addplot coordinates { (50,57.82) (100,61.34) (200,64.41)(400,66.25) (600,66.94)(800,66.62)};

\addplot coordinates { (50,45.49) (100,48.20) (200,50.77)(400,52.18) (600,53.23)(800,52.17)};

\addlegendentry{Proposed late fusion}

\addlegendentry{SIFT}

\addlegendentry{FREAK}

 \end{axis}
   \end{tikzpicture}
   \captionof{figure}{MAP as a function of codebook size (Corel-1500).}\label{fig:6}
\end{center}

\hfill \break
\hfill \break
The comparison results results presented in Fig. \ref{fig:3} indicate that the best MAP is obtained from the proposed last fusion on a codebook of size 600 with a value of 72.59 \%. The best MAP obtained by SIFT and FREAK is 66.94 \% and 53.23 \%, respectively. Table \ref{tab:4} presents a comparison of precision and recall values.

\begin{table}[!h]
\small
\centering
\caption{Comparison of precision and recall.}
\begin{tabular}
{|p{2.5cm}|>{\centering}p{2.3cm}|>{\centering}p{2.5cm}|>{\centering}p{2.4cm}|}
\hline
 Image retrieval performance  & Proposed late fusion & WA SIFT-SURF \cite{ali2016novel} & SQ + Spatiogram \cite{zeng2016image}
\tabularnewline  \hline
 \textbf{Precision} & \textbf{72.60} & 68.05 & 63.95  \tabularnewline \hline
\textbf{Recall} & \textbf{14.52} & 13.61 & 12.79   \tabularnewline \hline
\end{tabular}
\label{tab:4}
\end{table}

The proposed framework based on late fusion of FREAK and SIFT provides a better retrieval performance with higher values of precision and recall than the existing research \cite{zeng2016image,ali2016novel}.
\begin{figure}[!h]
\centering
    \includegraphics[width=0.90\textwidth, height=4cm]{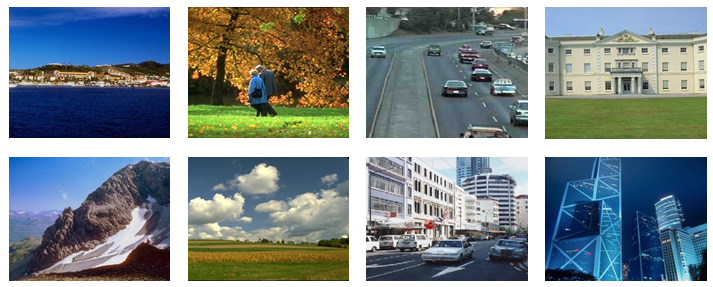}
      \caption{Randomly selected images from each class of OT-Scene image dataset \cite{oliva2001modeling}.}\label{ot8}
\end{figure}

\subsection{Performance Evaluation using Oliva and Torralba (OT-Scene)}

There are 08 classes in OT-Scene image dataset that contains a total of 2688 images. Fig. \ref{ot8} represents a sample of randomly selected images from each class of the OT-Scene image dataset. We evaluated the proposed framework using  OT-Scene benchmark and compared the results with state-of-the-art CBIR methods \cite{walia2014fusion,das2015multi}. Fig. \ref{fig:6} presents a comparison of mean precision as a function of codebook size using different evaluation parameters. Table \ref{tab:5} represents the comparison of MAP values.

\begin{center}
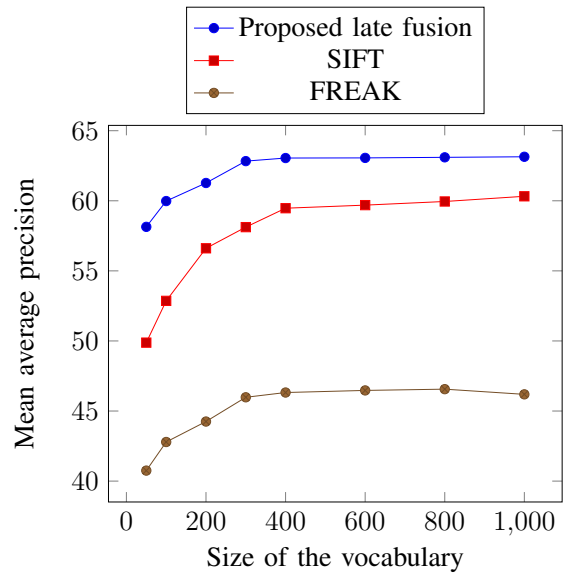

\begin{tikzpicture}[scale=0.88]

\pgfplotsset{every axis legend/.append style={
		at={(0.5,1.03)},
		anchor=south}}
\begin{axis}[legend columns=1,ylabel=Mean average precision,xlabel=Size of the vocabulary]
\addplot coordinates { (50,58.14) (100,59.98)  (200,61.27)(300,62.83) (400,63.05)(600,63.06)(800,63.1)(1000,63.14)};

\addplot coordinates { (50,49.88) (100,52.86)  (200,56.61)(300,58.12) (400,59.47)(600,59.69)(800,59.95)(1000,60.32)};

\addplot coordinates { (50,40.75) (100,42.79)  (200,44.25)(300,45.98) (400,46.32)(600,46.47)(800,46.56)(1000,46.19)};

\addlegendentry{Proposed late fusion}

\addlegendentry{SIFT}

\addlegendentry{FREAK}

 \end{axis}
   \end{tikzpicture}
   \captionof{figure}{MAP as a function of codebook size (OT-Scene).}\label{fig:6}
\end{center}

\begin{table}[!h]
\small
\centering
\caption{Mean precision comparison.}
\begin{tabular}
{|p{2.4cm}|>{\centering}p{2.2cm}|>{\centering}p{2.1cm}|>{\centering}p{2.0cm}|}
\hline
 Image retrieval performance& Proposed late fusion & Morphology based IR \cite{das2015multi} & Min Max Fusion \cite{walia2014fusion}
\tabularnewline  \hline
 \textbf{MAP} & \textbf{63.14} & 60.7 & 51.04
 \tabularnewline \hline

\end{tabular}
\label{tab:5}
\end{table}
The results and comparisons presented in  Fig. \ref{fig:6} and Table \ref{tab:5} indicate that the proposed late fusion of FREAK and SIFT provides a better retrieval performance than existing research \cite{walia2014fusion,das2015multi}.
\section{Conclusion and Future Directions}
\label{sec:5}

The research scope of this paper is based on the late fusion/visual word integration of binary and local descriptor for an effective image retrieval. Keeping in view, the good recognition ability and efficient computation time, we selected FREAK as the binary descriptor. SIFT is selected as the local feature as it is robust to change in translation, scaling, rotation, and small distortions. The image is represented in the form of late fusion of FREAK and SIFT.  The proposed research is based on the BoVW framework and classification is performed by using SVM. Testing is performed by varying the size of codebook, to sort out the best image retrieval precision. The results obtained from the proposed research are compared with state-of-the-art CBIR research.  The late fusion of FREAK and SIFT outperforms several CBIR methods. In future, we will evaluate our proposed work on real world retrieval problem by using a pre-trained CNN model.

\section*{Acknowledgments}
The authors would like to thank Professor James Z. Wang, Pennsylvania State University, USA for sharing Corel image benchmark that is used for the evaluation of the proposed research.


\begin{thebibliography}{}
\providecommand{\url}[1]{#1}
\csname url@samestyle\endcsname
\providecommand{\newblock}{\relax}
\providecommand{\bibinfo}[2]{#2}
\providecommand{\BIBentrySTDinterwordspacing}{\spaceskip=0pt\relax}
\providecommand{\BIBentryALTinterwordstretchfactor}{4}
\providecommand{\BIBentryALTinterwordspacing}{\spaceskip=\fontdimen2\font plus
\BIBentryALTinterwordstretchfactor\fontdimen3\font minus
  \fontdimen4\font\relax}
\providecommand{\BIBforeignlanguage}[2]{{%
\expandafter\ifx\csname l@#1\endcsname\relax
\typeout{** WARNING: IEEEtran.bst: No hyphenation pattern has been}%
\typeout{** loaded for the language `#1'. Using the pattern for}%
\typeout{** the default language instead.}%
\else
\language=\csname l@#1\endcsname
\fi
#2}}
\providecommand{\BIBdecl}{\relax}
\BIBdecl

\end{thebibliography}


\begin{thebibliography}{10}
\bibitem{dharani2013survey}
T.~Dharani and I.~L. Aroquiaraj, ``A survey on content based image retrieval,''
  in \emph{Pattern Recognition, Informatics and Mobile Engineering (PRIME),
  2013 International Conference on}.\hskip 1em plus 0.5em minus 0.4em\relax
  IEEE, 2013, pp. 485--490.

\bibitem{alzu2015semantic}
A.~Alzu'bi, A.~Amira, and N.~Ramzan, ``Semantic content-based image
  retrieval: A comprehensive study,'' \emph{Journal of Visual Communication and
  Image Representation}, vol.~32, pp. 20--54, 2015.

\bibitem{kumar2013content}
A.~Kumar, J.~Kim, W.~Cai, M.~Fulham, and D.~Feng, ``Content-based medical image
  retrieval: a survey of applications to multidimensional and multimodality
  data,'' \emph{Journal of digital imaging}, vol.~26, no.~6, pp. 1025--1039,
  2013.

\bibitem{rashedi2013simultaneous}
E.~Rashedi, H.~Nezamabadi-Pour, and S.~Saryazdi, ``A simultaneous feature
  adaptation and feature selection method for content-based image retrieval
  systems,'' \emph{Knowledge-Based Systems}, vol.~39, pp. 85--94, 2013.

\bibitem{zhang2012review}
D.~Zhang, M.~M. Islam, and G.~Lu, ``A review on automatic image annotation
  techniques,'' \emph{Pattern Recognition}, vol.~45, no.~1, pp. 346--362, 2012.

\bibitem{ali2016image}
N.~Ali, K.~B. Bajwa, R.~Sablatnig, and Z.~Mehmood, ``Image retrieval by
  addition of spatial information based on histograms of triangular regions,''
  \emph{Computers \& Electrical Engineering}, vol.~54, pp. 539--550, 2016.

\bibitem{mukherjee2015comparative}
D.~Mukherjee, Q.~J. Wu, and G.~Wang, ``A comparative experimental study of
  image feature detectors and descriptors,'' \emph{Machine Vision and
  Applications}, vol.~26, no.~4, pp. 443--466, 2015.

\bibitem{ashraf2014novel}
R.~Ashraf, T.~Mahmood, A.~Irtaza, and K.~Bajwa, ``A novel approach for the
  gender classification through trained neural networks,'' \emph{J. Basic Appl.
  Sci. Res}, vol.~4, pp. 136--144, 2014.

\bibitem{ashraf2015content}
R.~Ashraf, K.~Bashir, A.~Irtaza, and M.~T. Mahmood, ``Content based image
  retrieval using embedded neural networks with bandletized regions,''
  \emph{Entropy}, vol.~17, no.~6, pp. 3552--3580, 2015.

\bibitem{ashraf2016content}
R.~Ashraf, K.~Bashir, and T.~Mahmood, ``Content-based image retrieval by
  exploring bandletized regions through support vector machines,''
  \emph{Journal of Information Science and Engineering}, vol.~32, no.~2, pp.
  245--269, 2016.

\bibitem{mehmood2016novel}
Z.~Mehmood, S.~M. Anwar, N.~Ali, H.~A. Habib, and M.~Rashid, ``A novel image
  retrieval based on a combination of local and global histograms of visual
  words,'' \emph{Mathematical Problems in Engineering}, vol. 2016, 2016.

\bibitem{zhang2004review}
D.~Zhang and G.~Lu, ``Review of shape representation and description
  techniques,'' \emph{Pattern recognition}, vol.~37, no.~1, pp. 1--19, 2004.

\bibitem{lowe2004distinctive}
D.~G. Lowe, ``Distinctive image features from scale-invariant keypoints,''
  \emph{International journal of computer vision}, vol.~60, no.~2, pp. 91--110,
  2004.

\bibitem{bay2006surf}
H.~Bay, T.~Tuytelaars, and L.~Van~Gool, ``Surf: Speeded up robust features,''
  in \emph{Computer vision--ECCV 2006}.\hskip 1em plus 0.5em minus 0.4em\relax
  Springer, 2006, pp. 404--417.

\bibitem{dalal2005histograms}
N.~Dalal and B.~Triggs, ``Histograms of oriented gradients for human
  detection,'' in \emph{Computer Vision and Pattern Recognition, 2005. CVPR
  2005. IEEE Computer Society Conference on}, vol.~1.\hskip 1em plus 0.5em
  minus 0.4em\relax IEEE, 2005, pp. 886--893.

\bibitem{krajnik2015image}
T.~Krajnik, P.~Crist{\'o}foris, M.~Nitsche, K.~Kusumam, and T.~Duckett, ``Image
  features and seasons revisited,'' in \emph{Mobile Robots (ECMR), 2015
  European Conference on}.\hskip 1em plus 0.5em minus 0.4em\relax IEEE, 2015,
  pp. 1--7.

\bibitem{figat2014performance}
J.~Figat, T.~Kornuta, and W.~Kasprzak, ``Performance evaluation of binary
  descriptors of local features,'' in \emph{Computer Vision and
  Graphics}.\hskip 1em plus 0.5em minus 0.4em\relax Springer, 2014, pp.
  187--194.

\bibitem{leutenegger2011brisk}
S.~Leutenegger, M.~Chli, and R.~Y. Siegwart, ``Brisk: Binary robust invariant
  scalable keypoints,'' in \emph{Computer Vision (ICCV), 2011 IEEE
  International Conference on}.\hskip 1em plus 0.5em minus 0.4em\relax IEEE,
  2011, pp. 2548--2555.

\bibitem{alahi2012freak}
A.~Alahi, R.~Ortiz, and P.~Vandergheynst, ``Freak: Fast retina keypoint,'' in
  \emph{Computer Vision and Pattern Recognition (CVPR), 2012 IEEE Conference
  on}.\hskip 1em plus 0.5em minus 0.4em\relax Ieee, 2012, pp. 510--517.

\bibitem{calonder2010brief}
M.~Calonder, V.~Lepetit, C.~Strecha, and P.~Fua, ``Brief: Binary robust
  independent elementary features,'' \emph{Computer Vision--ECCV 2010}, pp.
  778--792, 2010.

\bibitem{viswanathan2009features}
D.~G. Viswanathan, ``Features from accelerated segment test (fast),'' 2009.

\bibitem{yu2013feature}
J.~Yu, Z.~Qin, T.~Wan, and X.~Zhang, ``Feature integration analysis of
  bag-of-features model for image retrieval,'' \emph{Neurocomputing}, vol. 120,
  pp. 355--364, 2013.

\bibitem{yuan2011sift}
X.~Yuan, J.~Yu, Z.~Qin, and T.~Wan, ``A sift-lbp image retrieval model based on
  bag of features,'' in \emph{IEEE International Conference on Image
  Processing}, 2011.

\bibitem{valgren2010sift}
C.~Valgren and A.~J. Lilienthal, ``Sift, surf \& seasons: Appearance-based
  long-term localization in outdoor environments,'' \emph{Robotics and
  Autonomous Systems}, vol.~58, no.~2, pp. 149--156, 2010.

\bibitem{localglobal}
Z.~Mehmood, S.~M. Anwar, M.~Altaf, and N.~Ali, ``A novel image retrieval based
  on rectangular spatial histograms of visual words,'' \emph{Kuwait Journal of
  Science}, 2017.

\bibitem{wang2013spatial}
C.~Wang, B.~Zhang, Z.~Qin, and J.~Xiong, ``Spatial weighting for
  bag-of-features based image retrieval,'' in \emph{Integrated Uncertainty in
  Knowledge Modelling and Decision Making}.\hskip 1em plus 0.5em minus
  0.4em\relax Springer, 2013, pp. 91--100.

\bibitem{zheng2010image}
Y.-b. Zheng, X.-s. Huang, and S.-j. Feng, ``An image matching algorithm based
  on combination of sift and the rotation invariant lbp,'' \emph{Journal of
  computer-aided design \& computer graphics}, vol.~22, no.~2, pp. 286--292,
  2010.

\bibitem{kabbai2015image}
L.~Kabbai, A.~Azaza, M.~Abdellaoui, and A.~Douik, ``Image matching based on lbp
  and sift descriptor,'' in \emph{Systems, Signals \& Devices (SSD), 2015 12th
  International Multi-Conference on}.\hskip 1em plus 0.5em minus 0.4em\relax
  IEEE, 2015, pp. 1--6.

\bibitem{zeng2016image}
S.~Zeng, R.~Huang, H.~Wang, and Z.~Kang, ``Image retrieval using spatiograms of
  colors quantized by gaussian mixture models,'' \emph{Neurocomputing}, vol.
  171, pp. 673--684, 2016.

\bibitem{walia2014fusion}
E.~Walia and A.~Pal, ``Fusion framework for effective color image retrieval,''
  \emph{Journal of Visual Communication and Image Representation}, vol.~25,
  no.~6, pp. 1335--1348, 2014.

\bibitem{dubey2015rotation}
S.~R. Dubey, S.~K. Singh, and R.~K. Singh, ``Rotation and scale invariant
  hybrid image descriptor and retrieval,'' \emph{Computers \& Electrical
  Engineering}, vol.~46, pp. 288--302, 2015.

\bibitem{montazer2015improved}
G.~A. Montazer and D.~Giveki, ``An improved radial basis function neural
  network for object image retrieval,'' \emph{Neurocomputing}, vol. 168, pp.
  221--233, 2015.

\bibitem{wan2014deep}
J.~Wan, D.~Wang, S.~C.~H. Hoi, P.~Wu, J.~Zhu, Y.~Zhang, and J.~Li, ``Deep
  learning for content-based image retrieval: A comprehensive study,'' in
  \emph{Proceedings of the ACM International Conference on Multimedia}.\hskip
  1em plus 0.5em minus 0.4em\relax ACM, 2014, pp. 157--166.

\bibitem{ali2016novel}
N.~Ali, K.~B. Bajwa, R.~Sablatnig, S.~A. Chatzichristofis, Z.~Iqbal, M.~Rashid,
  and H.~A. Habib, ``A novel image retrieval based on visual words integration
  of sift and surf,'' \emph{PloS one}, vol.~11, no.~6, p. e0157428, 2016.

\bibitem{sivic2003video}
J.~Sivic and A.~Zisserman, ``Video google: A text retrieval approach to object
  matching in videos,'' in \emph{Computer Vision, 2003. Proceedings. Ninth IEEE
  International Conference on}.\hskip 1em plus 0.5em minus 0.4em\relax IEEE,
  2003, pp. 1470--1477.

\bibitem{wang2001simplicity}
J.~Z. Wang, J.~Li, and G.~Wiederhold, ``Simplicity: Semantics-sensitive
  integrated matching for picture libraries,'' \emph{Pattern Analysis and
  Machine Intelligence, IEEE Transactions on}, vol.~23, no.~9, pp. 947--963,
  2001.

\bibitem{li2008real}
J.~Li and J.~Z. Wang, ``Real-time computerized annotation of pictures,''
  \emph{Pattern Analysis and Machine Intelligence, IEEE Transactions on},
  vol.~30, no.~6, pp. 985--1002, 2008.

\bibitem{oliva2001modeling}
A.~Oliva and A.~Torralba, ``Modeling the shape of the scene: A holistic
  representation of the spatial envelope,'' \emph{International journal of
  computer vision}, vol.~42, no.~3, pp. 145--175, 2001.

\bibitem{nowak2006sampling}
E.~Nowak, F.~Jurie, and B.~Triggs, ``Sampling strategies for bag-of-features
  image classification,'' in \emph{Computer Vision--ECCV 2006}.\hskip 1em plus
  0.5em minus 0.4em\relax Springer, 2006, pp. 490--503.

\bibitem{csurka2004visual}
G.~Csurka, C.~Dance, L.~Fan, J.~Willamowski, and C.~Bray, ``Visual
  categorization with bags of keypoints,'' in \emph{Workshop on statistical
  learning in computer vision, ECCV}, vol.~1, no. 1-22.\hskip 1em plus 0.5em
  minus 0.4em\relax Prague, 2004, pp. 1--2.

\bibitem{tian2014feature}
X.~Tian, L.~Jiao, X.~Liu, and X.~Zhang, ``Feature integration of eodh and
  color-sift: Application to image retrieval based on codebook,'' \emph{Signal
  Processing: Image Communication}, vol.~29, no.~4, pp. 530--545, 2014.

\bibitem{das2015multi}
R.~Das, S.~Thepade, and S.~Ghosh, ``Multi technique amalgamation for enhanced
  information identification with content based image data,''
  \emph{SpringerPlus}, vol.~4, no.~1, pp. 1--26, 2015.



\end{thebibliography}
\end{document}